\documentclass[10pt,conference]{IEEEtran}
\IEEEoverridecommandlockouts
\usepackage{cite}
\usepackage{amsmath,amssymb,amsfonts}
\usepackage{algorithm,algorithmic}
\usepackage{graphicx}
\usepackage{textcomp}
\usepackage[dvipsnames]{xcolor}
\usepackage{dirtytalk}
\def\BibTeX{{\rm B\kern-.05em{\sc i\kern-.025em b}\kern-.08em
    T\kern-.1667em\lower.7ex\hbox{E}\kern-.125emX}}
\makeatletter
\newcommand{\linebreakand}{%
  \end{@IEEEauthorhalign}
  \hfill\mbox{}\par
  \mbox{}\hfill\begin{@IEEEauthorhalign}
}
\makeatother

\usepackage{booktabs,tabularx}
\usepackage{array}
\newcolumntype{Y}{>{\centering\arraybackslash}X}
\usepackage{mathtools}
\usepackage{colortbl}
\usepackage[hidelinks]{hyperref}
\usepackage{orcidlink}

\newcommand{\texthl}[2]{%
  \begingroup
  \setlength{\fboxsep}{1pt}%
  \colorbox{#1!25}{#2}%
  \endgroup
}

\begin{document}

\title{SurvKAN: A Fully Parametric Survival Model Based on Kolmogorov--Arnold Networks}


\author{
\IEEEauthorblockN{%
Marina Mastroleo\IEEEauthorrefmark{1}\IEEEauthorrefmark{2}$^{\text{\orcidlink{0009-0005-7480-4102}}}$, %
Alberto Archetti\IEEEauthorrefmark{1}\IEEEauthorrefmark{2}$^{\text{\orcidlink{0000-0003-3826-4645}}}$, %
Federico Mastroleo\IEEEauthorrefmark{3}\IEEEauthorrefmark{4}\IEEEauthorrefmark{5}$^{\text{\orcidlink{0000-0001-6580-6767}}}$, %
Matteo Matteucci\IEEEauthorrefmark{2}$^{\text{\orcidlink{0000-0002-8306-6739}}}$}
\IEEEauthorblockA{\IEEEauthorrefmark{2}%
Department of Electronics, Information, and Bioengineering, Politecnico di Milano, Milan, Italy}
\IEEEauthorblockA{\IEEEauthorrefmark{3}Department of Radiation Oncology, Mayo Clinic, Rochester, MN, United States}
\IEEEauthorblockA{\IEEEauthorrefmark{4}Division of Radiation Oncology, IEO, European Institute of Oncology, IRCCS, Milan, Italy}
\IEEEauthorblockA{\IEEEauthorrefmark{5}Department of Oncology and Hemato-Oncology, University of Milan, Milan, Italy}
\IEEEcompsocitemizethanks{\IEEEcompsocthanksitem\IEEEauthorrefmark{1}Equal contribution.}
}


\maketitle

\begin{abstract}
Accurate prediction of time-to-event outcomes is critical for clinical decision-making, treatment planning, and resource allocation in modern healthcare. While classical survival models such as Cox remain widely adopted in standard practice, they rely on restrictive assumptions, including linear covariate relationships and proportional hazards over time, that often fail to capture real-world clinical dynamics. Recent deep learning approaches like DeepSurv and DeepHit offer improved expressivity but sacrifice interpretability, limiting clinical adoption where trust and transparency are paramount. Hybrid models incorporating Kolmogorov--Arnold Networks (KANs), such as CoxKAN, have begun to address this trade-off but remain constrained by the semi-parametric Cox framework. In this work we introduce SurvKAN, a fully parametric, time-continuous survival model based on KAN architectures that eliminates the proportional hazards constraint. SurvKAN treats time as an explicit input to a KAN that directly predicts the log-hazard function, enabling end-to-end training on the full survival likelihood. Our architecture preserves interpretability through learnable univariate functions that indicate how individual features influence risk over time. Extensive experiments on standard survival benchmarks demonstrate that SurvKAN achieves competitive or superior performance compared to classical and state-of-the-art baselines across concordance and calibration metrics. Additionally, interpretability analyses reveal clinically meaningful patterns that align with medical domain knowledge.
\end{abstract}


\section{Introduction}

The ability to predict when critical events will occur is fundamental to modern data-driven medicine. Accurate time-to-event predictions---whether for mortality, disease relapse, tumor recurrence, organ failure, or adverse drug reactions---directly impact treatment planning, resource allocation, and clinical decision-making in safety-critical environments.
Survival analysis addresses this challenge by leveraging statistical and machine learning techniques to estimate event occurrence probabilities over time~\cite{kleinbaum1996survival,wang2019machine}. Formally, given a set of patient features $\mathbf{x}$, a survival model estimates the survival function ${S(t \mid \mathbf{x})=P(T>t \mid \mathbf{x})}$, representing the probability that the event of interest has not occurred by time $t$. While traditionally focused on negative outcomes (mortality, disease progression, metastasis, etc.), this framework extends naturally to positive events such as recovery, remission, or hospital discharge. Beyond healthcare, survival analysis has proven valuable across diverse domains including reliability engineering, customer churn prediction, lifetime value estimation, and e-commerce conversion timing~\cite{liu2012survival}.

As healthcare data collection has evolved to encompass multimodal, high-dimensional, and increasingly complex patient records, classical survival models have begun to show fundamental limitations~\cite{wiegrebe2024deep}. Traditional approaches such as the Cox model~\cite{cox1972regression} rely on restrictive assumptions---notably linear relationships between covariates and risk, and proportional hazards (PH) over time---that often fail to capture real-world clinical dynamics. To overcome these constraints, recent work has incorporated modern machine learning techniques into survival modeling~\cite{kvamme2019time,kvamme2021continuous}. Neural network-based approaches such as DeepSurv~\cite{katzman2018deepsurv} and DeepHit~\cite{lee2018deephit} have demonstrated improved expressivity, but often at the cost of interpretability. This trade-off poses a significant barrier to clinical adoption: physicians operating under a \say{trust only what can be understood} principle are understandably reluctant to base life-critical decisions on opaque black-box models. Clinical deployment demands models whose internal reasoning can be examined, verified, and medically justified~\cite{archetti2025fpboost}.

The recent emergence of Kolmogorov--Arnold Networks (KANs)~\cite{liu2024kan,somvanshi2025survey} offers a promising resolution to this expressivity-interpretability dilemma. KANs fundamentally invert the architecture of traditional multi-layer perceptrons (MLPs). While MLPs place fixed activation functions on nodes and learnable parameters on edges, KANs place learnable activation functions (typically parameterized as spline functions) on edges while nodes simply perform summation. This architectural choice enables KANs to represent complex non-linear relationships with fewer parameters than comparable MLPs, while maintaining transparency: univariate transformations on edges can be directly visualized and interpreted. Furthermore, KAN architectures support systematic pruning and symbolic regression, allowing networks to be simplified into simple and interpretable mathematical expressions. In survival analysis, a recent KAN-based approach, CoxKAN~\cite{knottenbelt2025coxkan}, has demonstrated this potential by replacing linear predictors with KAN architectures while retaining the semi-parametric Cox framework. However, this hybrid model remains constrained by the PH assumption, limiting its ability to capture complex temporal dynamics where risk patterns evolve non-proportionally over time.

To address these issues, we introduce SurvKAN, a fully parametric, time-continuous survival model based on KAN architectures that eliminates the PH assumption. SurvKAN treats time as an explicit input to a single-output KAN that directly predicts the log-hazard function, from which patient-specific survival curves are reconstructed. This design offers two key advantages over existing approaches. First, as an end-to-end differentiable parametric model with no restrictive assumptions about temporal survival patterns, SurvKAN can be trained directly on the full survival likelihood, enabling it to learn arbitrary time-varying risk relationships. Second, the KAN-based architecture preserves interpretability: clinicians can examine how individual features influence risk over time, extracting medically meaningful patterns from the learned univariate functions.

We validate SurvKAN through extensive experiments on standard survival analysis benchmarks, demonstrating competitive or superior performance compared to both classical and state-of-the-art deep learning baselines across multiple evaluation metrics including concordance index and integrated Brier score. Additionally, we provide detailed interpretability analyses showing how SurvKAN captures clinically principled time-varying patterns, with feature effects that align with medical domain knowledge. 


\section{Related Work}
\label{sec:related}

\subsection{Survival Analysis}

Survival analysis aims to estimate the survival function $S(t \mid \mathbf{x}) = P(T > t \mid \mathbf{x})$, the probability that an individual with covariates $\mathbf{x} \in \mathbb{R}^d$ survives beyond time $t$~\cite{kleinbaum1996survival,wang2019machine}. A key challenge is right-censoring, where the event is not observed during follow-up and only a lower bound on the event time is known. Survival datasets consist of $N$ independent triplets $(\mathbf{x}_i, \delta_i, t_i)$, where $\delta_i \in \{0,1\}$ indicates whether the event occurred ($\delta_i = 1$) or was censored ($\delta_i = 0$), and $t_i$ denotes the observed time. Models may estimate the survival function directly or via the hazard function ${h(t \mid \mathbf{x}) = \lim_{\Delta t \rightarrow 0} P(t < T < t+\Delta t \mid T > t, \mathbf{x})/\Delta t}$, representing the instantaneous event rate. The cumulative hazard $H(t \mid \mathbf{x}) = \int_{0}^t h(u \mid \mathbf{x}) \, du$ relates to the survival function through $S(t \mid \mathbf{x}) = \exp(-H(t \mid \mathbf{x}))$, allowing models to estimate any of these quantities and deterministically reconstruct the others.

Survival models can be categorized into three types based on how the survival function is modeled~\cite{wang2019machine,wiegrebe2024deep}. Non-parametric approaches make no assumptions about the distributional form of survival times, relying instead on empirical estimates. Classical methods such as Kaplan-Meier, Nelson-Aalen, and Life Tables~\cite{kleinbaum1996survival} fall into this category. Modern non-parametric methods include tree-based models like Survival Trees and Random Survival Forests (RSF)~\cite{ishwaran2008random}, which apply recursive partitioning (CART) to stratify the data before computing classical estimators~\cite{archetti2023federated,archetti2023scaling}.

Semi-parametric models are based on the Cox proportional hazards formulation~\cite{cox1972regression} (CoxPH), which factorizes the hazard as $h(t \mid \mathbf{x}) = h_0(t) \exp(h_r(\mathbf{x}))$, where $h_0(t)$ is a time-dependent baseline hazard shared across the population and $h_r(\mathbf{x})$ is a covariate-dependent risk function. The classical Cox model assumes linear risk $h_r(\mathbf{x}) = \boldsymbol{\beta}^\top \mathbf{x}$ and optimizes the partial log-likelihood, which depends only on event ordering and eliminates the baseline hazard term. Extensions include DeepSurv~\cite{katzman2018deepsurv}, which replaces the linear risk with a neural network, and CoxBoost or XGBoost~\cite{chen2015xgboost}, which employ gradient-boosted trees.

Fully parametric models directly model the distribution of survival times as a function of covariates. Traditional approaches assume parametric families (exponential, Weibull, log-normal, Gompertz), with Accelerated Failure Time (AFT) models~\cite{wei1992accelerated} specifying $\log(T) = \boldsymbol{\beta}^\top \mathbf{x} + \epsilon$, where $\epsilon$ follows a known distribution. Modern neural network approaches achieve flexible parametric estimation without restrictive distributional assumptions~\cite{kvamme2019time,kvamme2021continuous,wiegrebe2024deep}. For instance, DeepHit~\cite{lee2018deephit} discretizes time into bins and frames survival prediction as multi-output classification, with the network outputting event probabilities for each bin.

\subsection{Kolmogorov--Arnold Networks}
\label{sec:kans}

Kolmogorov--Arnold Networks (KANs)~\cite{liu2024kan,somvanshi2025survey} represent a fundamental rethinking of neural network architecture inspired by the Kolmogorov--Arnold representation theorem, which states that any multivariate continuous function can be expressed as a finite composition of continuous univariate functions and addition. Unlike MLPs, which apply fixed activation functions at nodes with learnable weight matrices on edges, KANs apply learnable univariate activation functions on edges while nodes simply perform summation. Formally, for a KAN with layer widths $[n_0,n_1,\dots,n_L]$, the transformation from layer $l$ to $l+1$ is computed as ${x_{l+1,j} = \sum_{i=1}^{n_l} \phi_{l,j,i}(x_{l,i})}$ with $j$ from 1 to $n_{l+1}$ where each $\phi_{l,j,i}(\cdot):\mathbb{R}\to\mathbb{R}$ is a learnable univariate function. These functions are typically parameterized as B-splines with learnable coefficients. The full network is then $\mathrm{KAN}(\mathbf{x}) = (\Phi_{L-1}\circ\cdots\circ\Phi_0)(\mathbf{x})$, where $\Phi_l$ is the function matrix representing the transformation of the $l$-th layer. KANs demonstrate superior parameter efficiency, matching larger MLPs in accuracy with significantly fewer weights. Their architecture enables inherent interpretability through visualizable edge functions and supports symbolic discovery via automatic pruning and the extraction of closed-form mathematical expressions.

Within the field of survival analysis, CoxKAN~\cite{knottenbelt2025coxkan} represents the first application of KANs to time-to-event analysis, extending the Cox framework by replacing the traditional linear predictor with a KAN architecture. The model incorporates structured regularization (sparsity, smoothness, and entropy penalties) to enable automatic feature selection through pruning, followed by symbolic regression to extract interpretable mathematical formulas. While demonstrating superior performance to classical Cox models and competitive results with deep learning approaches, CoxKAN inherits the PH assumption, limiting its ability to capture temporal dynamics where hazard ratios vary over time.

\section{Method}

We propose SurvKAN, a KAN-based survival model addressing the limitations of existing methods by treating time as an explicit input feature alongside patient covariates, allowing the model to learn arbitrary time-varying hazard patterns directly from data. Unlike semi-parametric approaches that assume PH or discrete-time models that bin temporal information, SurvKAN models the continuous log-hazard function as the output of a KAN that takes both features $\mathbf{x}$ and time $t$ as input.

\subsection{SurvKAN Architecture}

Formally, SurvKAN models the log-hazard function as ${\log h(t \mid \mathbf{x}) = \mathrm{KAN}([\mathbf{x}, t])}$,
where $[\mathbf{x}, t] \in \mathbb{R}^{d+1}$ denotes the concatenation of patient features and time. The KAN follows the architecture described in Section~\ref{sec:kans}, with input dimension $d+1$, hidden layer of size $m$, and single output neuron (width configuration $[d+1, m, 1],\,m\in\{0,\dots,3\}$). Each edge function is parameterized as $\phi(x) = w_b \cdot b(x) + w_s \cdot \mathrm{spline}(x)$, where $w_b,w_s\in\mathbb{R}$ and $b(x)$ is a fixed base function (Identity or SiLU activation) and $\mathrm{spline}(x)$ is a quadratic B-spline. Time is normalized to $[0,1]$ during training for numerical stability. The survival function is then reconstructed from the predicted log-hazard. Since the KAN output is continuous in time but returns a single risk per evaluation, we approximate the cumulative hazard $H(t \mid \mathbf{x})$ via trapezoidal rule over $K$ uniformly spaced points in $[0,t]$.

\subsection{Training Procedure}

SurvKAN is trained by maximizing the full survival likelihood. The training objective minimizes the negative log-likelihood ${\mathcal{L}_{\text{NLL}} = -\frac{1}{N}\sum_{i=1}^N \left[\delta_i \log h(t_i \mid \mathbf{x}_i) - H(t_i \mid \mathbf{x}_i)\right]}$, where log-hazard values are clamped to $[-20, 20]$ to prevent numerical instability during exponentiation. To preserve the interpretability and sparsity properties of KANs from their original implementation~\cite{liu2024kan}, the likelihood is augmented with structured regularization that encourages prunable, smooth, and meaningful edge functions. The complete training objective is $\mathcal{L} = \mathcal{L}_{\text{NLL}} + \lambda\sum_{l,j,i}\Omega_{l,j,i}$ where $\Omega_{l,j,i}$ is the weighted sum of four regularization terms acting on each edge function $\phi_{l,j,i}$ in the network. Specifically, the first term $\|\phi_{l,j,i}\|_1$ penalizes the $L_1$ norm of edge outputs, enabling automatic feature selection by driving unimportant edges toward zero. The second term $H(\phi_{l,j,i})$ is an entropy penalty that prevents a few edges from dominating the network, ensuring balanced contributions. The third and fourth terms act on the B-spline coefficients $\mathbf{c}_{l,j,i}$: $\|\mathbf{c}_{l,j,i}\|_2$ penalizes coefficient magnitude, while $\|\Delta \mathbf{c}_{l,j,i}\|_2$ penalizes differences between adjacent coefficients to promote smoothness. The SurvKAN architecture is end-to-end differentiable and can therefore be trained using standard gradient-based optimizers via automatic differentiation, enabling seamless integration with modern deep learning frameworks~\cite{loshchilov2017decoupled}.

\subsection{Inference and Interpretability}

At inference time, SurvKAN predicts patient-specific survival curves by evaluating the trained model at a temporal grid $\{t_1, \ldots, t_M\}$ spanning the time horizon of interest. For each time point $t_j$, we compute the hazard $h(t_j \mid \mathbf{x}_{\text{new}})$ and apply trapezoidal integration to obtain the cumulative hazard $H(t_j \mid \mathbf{x}_{\text{new}})$, from which the survival probability follows as $S(t_j \mid \mathbf{x}_{\text{new}}) = \exp(-H(t_j \mid \mathbf{x}_{\text{new}}))$. Since the underlying KAN is a continuous function of time, finer temporal grids yield increasingly accurate approximations of the true survival curve, providing an inherent advantage with respect to discrete-time models with limited temporal resolution.

The interpretability of SurvKAN derives directly from its learned univariate edge functions $\phi_{l,j,i}(\cdot)$. By visualizing these learned functions, clinicians can examine the precise functional relationship between each covariate and the predicted log-hazard. For the time input, the edge function $\phi(\cdot, t)$ reveals how patient hazard evolves over time, identifying phases of elevated or diminished risk. For individual features $x_k$, the corresponding edge functions $\phi(x_k, \cdot)$ expose the corresponding non-linear marginal effects. Furthermore, the KAN framework supports post-hoc symbolic regression, where learned B-splines can be approximated by closed-form expressions (polynomials, exponentials, logarithms, etc.) to produce human-readable mathematical formulas. We showcase these interpretability capabilities in Section~\ref{sec:interpretability}.

\section{Experiments}

\begin{table*}[t]
\caption{C-index ($\uparrow$) mean $\pm$ 95\% confidence interval over 5 independent hold-out splits for each dataset. All values are scaled by a factor of 100 for readability. For each dataset, the best model is highlighted in \texthl{green}{\textbf{bold}}, while the second-best is \texthl{yellow}{\underline{underlined}}.}
\label{tab:cindex}
\begin{center}
\begin{small}
\begin{sc}
\begin{tabularx}{\textwidth}{@{} l | Y Y Y Y Y Y Y | Y}
\toprule
Dataset & AFT & CoxPH & CoxKAN & DeepHit & DeepSurv & RSF & XGBoost & SurvKAN\\
\midrule
AIDS & $77.3\pm3.3$ & $78.7\pm3.7$ & \cellcolor{yellow!25}$\underline{78.8\pm5.2}$ & $74.1\pm8.1$ & $77.6\pm2.8$ & $78.2\pm6.8$ & $78.1\pm9.0$ & \cellcolor{green!25}$\mathbf{79.2\pm4.9}$\\
Breast Cancer & $58.6\pm17.2$ & $59.4\pm14.5$ & $60.0\pm12.6$ & $65.0\pm17.9$ & $62.7\pm13.5$ & \cellcolor{green!25}$\mathbf{65.6\pm13.1}$ & \cellcolor{yellow!25}$\underline{65.0\pm13.0}$ & $64.0\pm22.8$\\
Framingham & $71.1\pm5.0$ & $71.2\pm5.2$ & \cellcolor{yellow!25}$\underline{71.4\pm5.1}$ & $70.5\pm4.9$ & $70.9\pm4.3$ & $70.0\pm4.0$ & $70.9\pm4.8$ & \cellcolor{green!25}$\mathbf{71.4\pm4.6}$\\
GBSG2 & $67.5\pm4.5$ & $64.1\pm6.5$ & $67.0\pm8.1$ & $67.3\pm3.9$ & $66.9\pm5.2$ & $67.2\pm6.5$ & \cellcolor{yellow!25}$\underline{67.6\pm4.3}$ & \cellcolor{green!25}$\mathbf{68.1\pm5.7}$\\
METABRIC & $64.1\pm4.4$ & $63.6\pm2.1$ & \cellcolor{green!25}$\mathbf{64.9\pm3.2}$ & $64.5\pm3.4$ & $63.9\pm5.2$ & $63.4\pm4.3$ & $64.6\pm2.5$ & \cellcolor{yellow!25}$\underline{64.7\pm2.7}$\\
NWTCO & $71.0\pm4.8$ & $70.5\pm6.2$ & $71.6\pm4.2$ & $71.5\pm6.0$ & $71.0\pm5.8$ & \cellcolor{yellow!25}$\underline{71.9\pm4.2}$ & \cellcolor{green!25}$\mathbf{72.1\pm5.1}$ & $71.7\pm2.8$\\
PBC & $71.1\pm11.2$ & $72.1\pm12.3$ & $72.3\pm12.0$ & $73.3\pm12.2$ & $72.8\pm13.2$ & $72.8\pm11.8$ & \cellcolor{yellow!25}$\underline{73.5\pm12.5}$ & \cellcolor{green!25}$\mathbf{74.2\pm10.2}$\\
TCGA--BRCA & $65.4\pm6.3$ & \cellcolor{green!25}$\mathbf{72.0\pm10.1}$ & $70.8\pm14.1$ & $69.9\pm10.7$ & \cellcolor{yellow!25}$\underline{70.9\pm12.3}$ & $66.4\pm10.3$ & $70.7\pm11.7$ & $70.5\pm15.3$\\
Veterans & $70.5\pm7.5$ & $69.8\pm8.1$ & $65.0\pm21.8$ & $73.0\pm12.6$ & \cellcolor{green!25}$\mathbf{73.9\pm8.2}$ & $68.9\pm7.7$ & $69.7\pm9.7$ & \cellcolor{yellow!25}$\underline{73.0\pm13.1}$\\
WHAS500 & $72.5\pm4.4$ & $74.4\pm10.4$ & $75.2\pm10.7$ & $74.0\pm8.1$ & $74.3\pm9.7$ & \cellcolor{yellow!25}$\underline{76.1\pm8.3}$ & \cellcolor{green!25}$\mathbf{76.3\pm11.5}$ & $74.6\pm12.0$\\
\bottomrule
\end{tabularx}
\end{sc}
\end{small}
\end{center}
\end{table*}

\begin{table*}[t]
\caption{IBS ($\downarrow$) mean $\pm$ 95\% confidence interval over 5 independent hold-out splits for each dataset. All values are scaled by a factor of 100 for readability. For each dataset, the best model is highlighted in \texthl{green}{\textbf{bold}}, while the second-best is \texthl{yellow}{\underline{underlined}}.}
\label{tab:ibs}
\begin{center}
\begin{small}
\begin{sc}
\begin{tabularx}{\textwidth}{@{} l | Y Y Y Y Y Y Y | Y}
\toprule
Dataset & AFT & CoxPH & CoxKAN & DeepHit & DeepSurv & RSF & XGBoost & SurvKAN\\
\midrule
AIDS & $6.5\pm0.5$ & \cellcolor{green!25}$\mathbf{6.2\pm0.5}$ & \cellcolor{yellow!25}$\underline{6.2\pm0.5}$ & $6.5\pm0.5$ & $6.3\pm0.4$ & $6.5\pm0.4$ & $6.3\pm0.4$ & $6.3\pm0.4$\\
Breast Cancer & $20.3\pm8.8$ & $20.1\pm6.6$ & $18.7\pm3.6$ & \cellcolor{yellow!25}$\underline{17.6\pm5.5}$ & $19.4\pm13.2$ & \cellcolor{green!25}$\mathbf{15.6\pm1.5}$ & $20.5\pm6.3$ & $18.7\pm9.8$\\
Framingham & $13.3\pm1.0$ & $13.2\pm1.0$ & \cellcolor{green!25}$\mathbf{13.1\pm1.0}$ & $14.1\pm0.8$ & $13.2\pm0.9$ & $13.5\pm0.7$ & $13.2\pm0.9$ & \cellcolor{yellow!25}$\underline{13.1\pm0.8}$\\
GBSG2 & $18.5\pm2.2$ & $19.3\pm1.6$ & \cellcolor{yellow!25}$\underline{18.4\pm2.9}$ & $19.1\pm2.4$ & $18.8\pm2.0$ & $18.8\pm2.0$ & $18.4\pm1.5$ & \cellcolor{green!25}$\mathbf{18.3\pm2.6}$\\
METABRIC & $19.5\pm1.3$ & $19.3\pm0.9$ & $18.9\pm0.8$ & \cellcolor{yellow!25}$\underline{18.8\pm1.5}$ & $19.3\pm1.1$ & $19.2\pm1.1$ & $19.1\pm1.1$ & \cellcolor{green!25}$\mathbf{18.7\pm0.4}$\\
NWTCO & \cellcolor{green!25}$\mathbf{10.5\pm0.5}$ & $11.2\pm0.9$ & $10.6\pm0.6$ & $10.7\pm0.7$ & $10.8\pm0.6$ & $10.7\pm0.4$ & \cellcolor{yellow!25}$\underline{10.6\pm0.9}$ & $10.7\pm0.6$\\
PBC & $18.1\pm4.4$ & $17.5\pm3.5$ & $17.4\pm3.8$ & $17.1\pm3.9$ & $17.7\pm4.1$ & \cellcolor{yellow!25}$\underline{16.7\pm3.6}$ & $17.0\pm4.2$ & \cellcolor{green!25}$\mathbf{16.7\pm3.4}$\\
TCGA--BRCA & $11.4\pm2.4$ & \cellcolor{yellow!25}$\underline{10.2\pm1.8}$ & $10.5\pm1.9$ & $11.0\pm1.5$ & $10.4\pm2.5$ & \cellcolor{green!25}$\mathbf{10.0\pm1.6}$ & $10.6\pm1.9$ & $11.1\pm1.6$\\
Veterans & $16.9\pm6.2$ & $15.9\pm4.8$ & $17.6\pm7.8$ & $17.1\pm4.4$ & \cellcolor{green!25}$\mathbf{15.4\pm4.0}$ & $15.8\pm3.4$ & \cellcolor{yellow!25}$\underline{15.5\pm2.0}$ & $15.6\pm5.3$\\
WHAS500 & $17.8\pm2.9$ & $16.1\pm3.5$ & \cellcolor{green!25}$\mathbf{15.8\pm4.9}$ & $16.5\pm2.9$ & $16.5\pm3.4$ & $16.3\pm2.3$ & $16.4\pm5.1$ & \cellcolor{yellow!25}$\underline{15.9\pm4.8}$\\
\bottomrule
\end{tabularx}
\end{sc}
\end{small}
\end{center}
\end{table*}

\begin{table}[t]
\caption{Relative improvement (\%) of SurvKAN compared to the average baseline metric in each category.}
\label{tab:aggr}
\begin{center}
\begin{small}
\begin{sc}
\begin{tabularx}{\linewidth}{@{} l | Y Y}
\toprule
Baseline Type & C-Index (\%) & IBS (\%)\\
\midrule
Linear & \cellcolor{green!25}$+2.27$ & \cellcolor{green!25}$+2.65$\\
Tree-Based & \cellcolor{green!25}$+1.00$ & \cellcolor{green!25}$+0.19$\\
Neural-Based & \cellcolor{green!25}$+1.77$ & \cellcolor{green!25}$+3.01$\\
KAN-Based & \cellcolor{green!25}$+2.05$ & \cellcolor{green!25}$+1.45$\\
\midrule
Non-Parametric & \cellcolor{green!25}$+1.58$ & \cellcolor{red!25}$-1.38$\\
Semi-Parametric & \cellcolor{green!25}$+1.41$ & \cellcolor{green!25}$+1.87$\\
Fully Parametric & \cellcolor{green!25}$+1.19$ & \cellcolor{green!25}$+2.24$\\
\bottomrule
\end{tabularx}
\end{sc}
\end{small}
\end{center}
\end{table}

\subsection{Experimental Setup}


We evaluate our methods on ten publicly available survival datasets drawn from standard benchmarks across diverse medical domains. The datasets are sourced from scikit-survival~\cite{polsterl2020scikit} (AIDS, GBSG2, Veterans, WHAS500), SurvSet~\cite{drysdale2022survset} (Breast Cancer, Framingham, NWTCO, PBC), Flamby~\cite{ogier2022flamby} (TCGA--BRCA), and DeepSurv~\cite{katzman2018deepsurv} (METABRIC). 



Models are evaluated using two complementary metrics that assess different aspects of survival estimation: concordance index (C-Index)~\cite{harrell1996multivariable} and Integrated Brier Score (IBS)~\cite{glenn1950verification}. The C-Index measures the discriminative ability of a survival model by quantifying how well its predictions rank patients according to risk. The IBS, instead, measures calibration by computing the time-averaged mean squared error between predicted survival probabilities and observed outcomes. Together, these metrics provide a comprehensive assessment of both ranking performance and probabilistic accuracy~\cite{archetti2024bridging,lillelund2025stop}.

We conduct a rigorous evaluation across all dataset-model combinations using stratified 80-20 train-test splits to preserve event rate distributions, repeated over 5 independent runs with different random seeds. Each metric reported is averaged over all the independent seed executions. For each run, we perform hyperparameter optimization using Optuna~\cite{akiba2019optuna} with 100 trials and multi-objective optimization targeting two objectives: maximizing C-Index and minimizing IBS. The optimization employs the NSGA III sampler~\cite{deb2013evolutionary} to identify Pareto-optimal hyperparameter configurations. During optimization, candidate hyperparameters are evaluated via 5-fold stratified cross-validation on the training set. Neural network models are trained with early stopping using a further 15\% internal split derived from the training data. Once the best hyperparameters are identified, we train the final model on the full training set and evaluate its performance on the held-out test set.

Our experiments include seven models spanning classical, tree-based, neural network, and KAN-based approaches alongside SurvKAN. These models, described in Section~\ref{sec:related}, include AFT, CoxPH, CoxKAN, DeepHit, DeepSurv, RSF, and XGBoost.
Predictions from models with discrete outputs are linearly interpolated~\cite{archetti2023deep,archetti2024bridging}. Concerning SurvKAN’s numerical integration, we empirically set $K=50$, yielding a reasonable computational cost with diminishing returns for larger values.

\subsection{Results}

Tables~\ref{tab:cindex} and \ref{tab:ibs} present the concordance index and integrated Brier score results across all datasets and baseline models. SurvKAN demonstrates competitive performance, achieving either the best or second-best ranking on 6 out of 10 datasets for C-index (AIDS, Framingham, GBSG2, METABRIC, PBC, Veterans) and 5 out of 10 datasets for IBS (Framingham, GBSG2, METABRIC, PBC, WHAS500).

Table~\ref{tab:aggr} provides aggregate comparisons between SurvKAN and different model categories. The results reveal several trends. First, as expected, SurvKAN demonstrates substantial improvements over CoxPH, the only linear model, achieving $+2.27$\% in C-index and $+2.65$\% in IBS. Second, comparing against KAN-based methods, SurvKAN outperforms CoxKAN by $+2.05$\% in C-index and $+1.45$\% in IBS, demonstrating that eliminating the PH assumption enables more flexible temporal modeling. Third, SurvKAN shows a strong improvement over neural network baselines ($+1.77$\% C-index, $+3.01$\% IBS), suggesting that the structured inductive bias of KAN architectures can provide both expressivity and regularization advantages over standard MLPs in the survival setting. Improvements with respect to tree-based models are modest but still noticeable in terms of discrimination ($+1$\% in C-index).

The comparison between parametric formulations reveals interesting trade-offs. While SurvKAN shows modest improvements over other fully parametric models in C-index ($+1.19$\%), it demonstrates stronger gains in calibration ($+2.24$\% IBS improvement), indicating that the KAN-based hazard modeling produces more reliable probability estimates than discrete-time approaches like DeepHit. Against semi-parametric methods, SurvKAN achieves balanced improvements in both discrimination and calibration. The one exception appears in comparison to non-parametric methods (specifically RSF), where SurvKAN shows improved discrimination ($+1.58$\% C-index) but slightly degraded calibration ($-1.38$\% IBS). This suggests that the assumption of continuous hazard functions, while beneficial for learning smooth risk patterns, may introduce small systematic biases in probability estimation that ensemble methods like RSF avoid through their distribution-free nature.

Overall, the experimental results demonstrate that SurvKAN provides a convincing balance between predictive performance and model transparency by achieving competitive or superior results across diverse datasets and baselines.

\section{Interpretability}
\label{sec:interpretability}

\begin{figure}[t]
    \includegraphics[width=\linewidth]{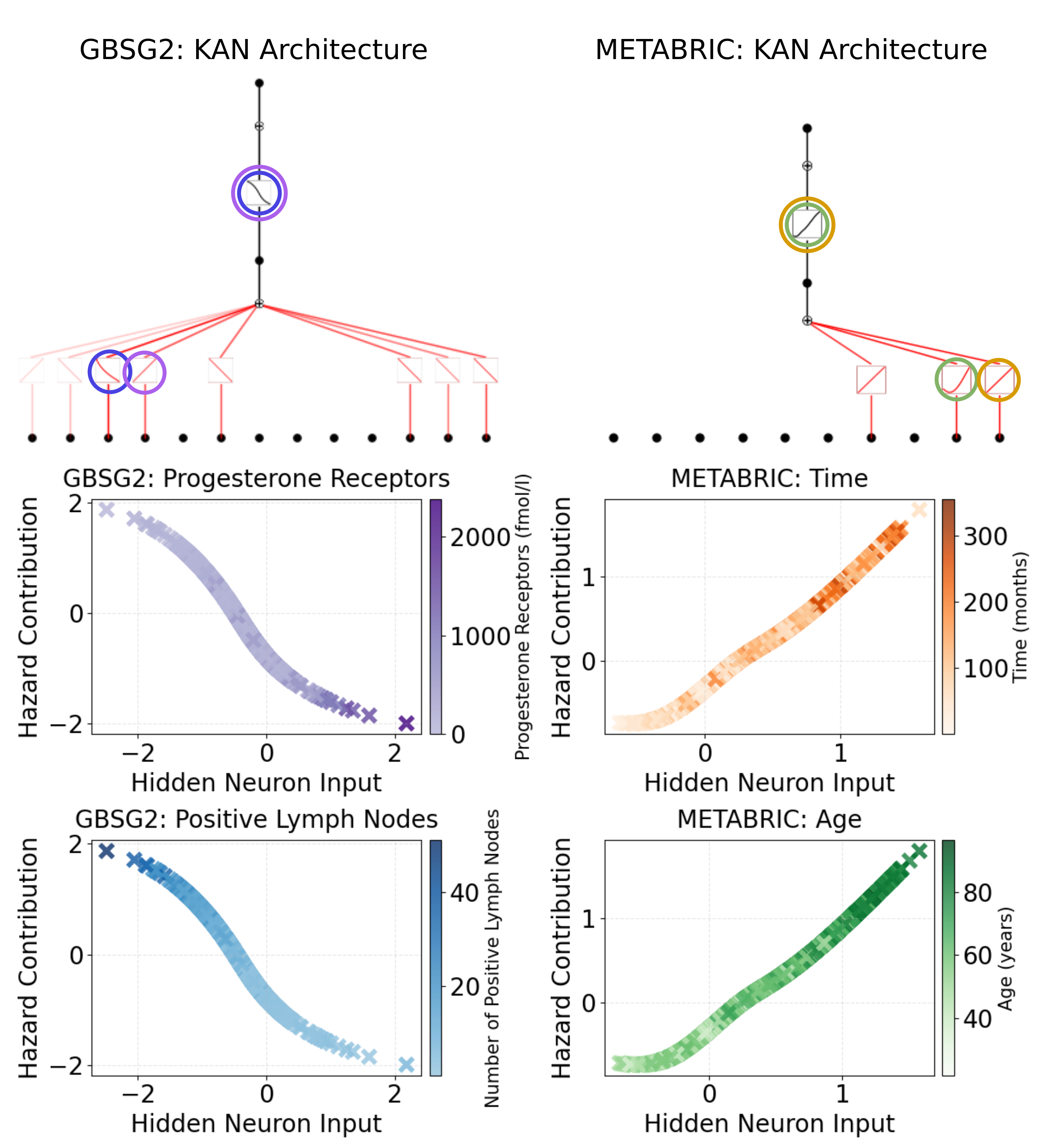}
    \caption{KAN architecture schema from inputs (bottom) to log-hazard (top) alongside feature-specific hazard contributions in the subsequent rows.}
    \label{fig:interp}
\end{figure}

In oncology, when patient lives are at stake, the opacity of black-box models creates an unacceptable risk by obscuring potentially unsafe correlations. SurvKAN addresses this by distilling the network into a compact symbolic expression via structural pruning~\cite{liu2024kan}. This process autonomously identifies clinically meaningful non-linear relationships and yields a transparent mathematical form that can be inspected and validated by clinicians. We evaluate the interpretability of SurvKAN on two breast cancer datasets with distinct clinical endpoints, GBSG2 and METABRIC. GBSG2 tracks recurrence-free survival~\cite{konstantinov2024}, for which SurvKAN prioritized tumor biology (lymph nodes, tumor grade) governing relapse. Conversely, METABRIC records overall survival~\cite{Thammasorn2024} and SurvKAN shifted to systemic factors like age and time, capturing frailty-dependent mortality. This confirms the capacity of our model to autonomously adapt feature selection to the specific prognostic drivers (biological versus demographic) of the target outcome. The interpretation results are summarized in Figure~\ref{fig:interp}. The left column displays results for the GBSG2 dataset, while the right column corresponds to METABRIC. The top row illustrates the pruned KAN architectures optimized for each dataset. Subsequent rows visualize feature-specific hazard contributions, detailing how individual variables influence model predictions.


\subsection{The GBSG2 Dataset}

Post-pruning, SurvKAN yields the following symbolic log-hazard function:
\begin{small}
\begin{equation*}
\begin{aligned}
\log(h(\mathbf{x} \mid t)) &= - 0.05 (\text{Age}) + 0.68 (\text{Time}) + 0.06 (\text{Tumor size})  \\
&\quad + 0.38 (\text{Tumor Grade II}) + 0.3 (\text{Tumor Grade III}) \\
&\quad - 0.3 (\text{Progesterone Receptors}) \\
&\quad + 0.37 (\text{No Hormone Therapy}) \\
&\quad + 0.65 \sqrt{3.07 (\text{Positive Lymph Nodes}) + 2.88} \\
&\quad - 2.01.
\end{aligned}
\end{equation*}
\end{small}

The discovered formula reveals a hybrid risk structure by applying a square root transformation to the number of \textit{Positive Lymph Nodes}. This captures a clinical saturation effect: while the detection of the first few nodes causes a drastic spike in mortality risk signaling metastasis, the marginal impact of additional nodes diminishes beyond a threshold of approximately 20, as shown in Figure~\ref{fig:interp} (\textcolor{Cerulean}{blue}). Concerning linear factors, \textit{Tumor Grade} and \textit{Tumor Size} are correctly identified as risk enhancers. Conversely, \textit{Progesterone Receptors} and \textit{Age} function as protective factors. The negative coefficient for \textit{Age} likely indicates that younger patients present with biologically more aggressive tumor phenotypes compared to the slower-growing cancers common in post-menopausal women. Similarly, the protective effect of \textit{Progesterone Receptors} (Figure~\ref{fig:interp}, \textcolor{DarkOrchid}{orchid}) aligns with the presence of well-differentiated, hormone-sensitive tumors that respond favorably to endocrine therapies. Finally, the variable \textit{Time} exhibits a positive linear coefficient, implying a multiplicative (exponential) increase in hazard over time, consistent with the biological reality of accelerating recurrence risk.

\subsection{The METABRIC Dataset}

On the METABRIC dataset, the pruning process yielded a more concise symbolic representation:

\begin{small}
\begin{equation*}
\begin{aligned}
\log(h(\mathbf{x} \mid t)) &= 1.73 (\text{Time}) + 0.26 (\text{Chemotherapy}) \\
& \quad + 1.29 \sin{\left(0.61 (\text{Age}) - 0.78 \right)} + 0.28.
\end{aligned}
\end{equation*}
\end{small}

This equation highlights several nonlinear biological dynamics that linear models often oversimplify. Instead of assuming risk rises linearly with \textit{Age}, SurvKAN discovered that breast cancer risk increases for middle-aged women but plateaus for the elderly (within the age range of this patient cohort, as shown in Figure~\ref{fig:interp}, \textcolor{Green}{green}). 
This effectively captures the \say{ceiling effect} often attributed to competing risks, where elderly patients are increasingly likely to succumb to comorbidities rather than the cancer itself. Regarding treatment, the model assigns a positive coefficient to \textit{Chemotherapy}. Although this seems counter-intuitive for a therapy, since chemotherapy is preferentially administered to high-risk patients, the variable serves as a marker for the underlying severity of the disease rather than the effect of the treatment itself. This interpretation is associative rather than causal, as treatment assignment is confounded with disease severity. Furthermore, the strong positive coefficient for \textit{Time} validates the ability of the model to process time as a continuous input (Figure~\ref{fig:interp}, \textcolor{orange}{orange}). Thus, by learning a monotonic relation between early timepoints to long-term survival, SurvKAN successfully reconstructs biologically sound hazard patterns.



\section{Conclusion}

We introduced SurvKAN, a fully parametric, time-continuous survival model based on Kolmogorov--Arnold Networks. By treating time as an explicit input and directly modeling the log-hazard function through learnable univariate transformations, SurvKAN addresses a fundamental limitation of existing approaches: the trade-off between expressive power and interpretability. Our extensive evaluation demonstrates that SurvKAN achieves competitive or superior performance compared to classical and state-of-the-art baselines, revealing systematic improvements over neural, tree-based, and semi-parametric models. Beyond predictive accuracy, SurvKAN enables direct visualization of how individual features influence risk over time, providing the interpretability necessary for clinical deployment in safety-critical settings.

\bibliographystyle{IEEEtran}
\bibliography{references}

\end{document}